\documentclass[11pt]{article}

\PassOptionsToPackage{numbers, compress}{natbib}

\usepackage{natbib}
\setcitestyle{authoryear,open={(},close={)}}

\usepackage{amsmath}
\usepackage{amsthm}
\usepackage{amssymb}
\numberwithin{equation}{section}

\usepackage{natbib}
\usepackage{hyperref}
\usepackage{url} 
\usepackage{booktabs}       
\usepackage{amsfonts}       
\usepackage{nicefrac}       
\usepackage{microtype}      
\usepackage{xcolor}         
\usepackage{graphicx}
\usepackage{caption}
\usepackage{bbm}
\usepackage{algorithm}
\usepackage{algpseudocode}
\usepackage{amsmath,amssymb}
\usepackage{enumitem}
\usepackage{comment} 
\usepackage{hyperref}
\hypersetup{colorlinks,
	linkcolor=blue,
	citecolor=blue,
	urlcolor=magenta,
	linktocpage,
	plainpages=false}
\usepackage{natbib}
\usepackage{bm}

\usepackage{thmtools}
\usepackage{thm-restate}

\usepackage{times}

\usepackage{amssymb}
\usepackage{amsmath}
\usepackage{amsfonts}
\usepackage{amsthm}

\usepackage[utf8]{inputenc} 
\usepackage[T1]{fontenc}    
\usepackage{url}            
\usepackage{booktabs}       
\usepackage{amsfonts}       
\usepackage{nicefrac}       
\usepackage{microtype}      
  \usepackage{mathtools}

\usepackage{tikz}
\usepackage{pgfplots}
\usetikzlibrary{pgfplots.groupplots}

\usepackage{caption}
\usepackage[bottom,hang,flushmargin]{footmisc}

\newcommand{\tsn}[1]{{\left\vert\kern-0.25ex\left\vert\kern-0.25ex\left\vert #1 
    \right\vert\kern-0.25ex\right\vert\kern-0.25ex\right\vert}}

\newenvironment{itemize*}%
{\begin{itemize}[leftmargin=*,topsep=0pt]%
		\setlength{\itemsep}{0pt}%
		\setlength{\parskip}{0pt}}%
	{\end{itemize}}
\newenvironment{enumerate*}%
{\begin{enumerate}[leftmargin=*,topsep=0pt]%
		\setlength{\itemsep}{0pt}%
		\setlength{\parskip}{0pt}}%
	{\end{enumerate}}

\allowdisplaybreaks

\newtheorem{theorem}{Theorem}[section]

\newtheorem{claim}[theorem]{Claim}
\newtheorem{assumption}{Assumption}[section]

\newtheorem{lemma}[theorem]{Lemma}
\newtheorem{corollary}[theorem]{Corollary}
\newtheorem{proposition}[theorem]{Proposition}

\newtheorem{remark}{Remark}

\newcommand{\vct}{\boldsymbol }

\def\R{\mathbb{R}}

\def\cA{\mathcal{A}}

\def\cD{\mathcal{D}}

\def\cN{\mathcal{N}}

\newtheorem{prop}{Proposition}[section]

\newtheorem{fact}{Fact}[section]

\def\approxcorrect{\cmark\kern-1.4ex\raisebox{.30ex}{$\xmark$}}

\newcommand{\idxn}[1][]{\ifthenelse{\equal{#1}{}}{\mathbb{INDQ}_n}{\mathbb{INDQ}_{#1}}}

\newcommand{\beq}{\begin{equation}}

\newcommand{\eeq}{\end{equation}}

\newcommand{\bx}{{\bm{x}}}

\newcommand{\bw}{{\bm{w}}}

\newcommand{\bE}{{\bm{E}}}

\newcommand{\bT}{{\bm{T}}}

\newcommand{\ba}{{\bm{a}}}
\newcommand{\bu}{{\bm{u}}}
\newcommand{\bv}{{\bm{v}}}
\newcommand{\bz}{{\bm{z}}}

\newcommand{\be}{{\bm{e}}}

\newcommand{\w}{\vct{w}}

\newcommand{\bb}{\bm{b}}

\definecolor{emmanuel}{RGB}{255,127,0}

\newcommand{\E}{\operatorname{\mathbb{E}}}

\def \endprf{\hfill {\vrule height6pt width6pt depth0pt}\medskip}

\usepackage{footnote}
\usepackage{authblk}
\usepackage{fullpage} 

\ifdefined\usebigfont

\usepackage{times}
\usepackage[fontsize=13pt]{scrextend}
\usepackage[letterpaper,left=1.56in,right=1.56in,top=1.71in,bottom=1.77in]{geometry}
\else
\fi

\title{Reconstructing Training Data from Model Gradient, Provably}
\author{
	Zihan Wang\textsuperscript{1}, Jason D. Lee\textsuperscript{2}, and Qi Lei\textsuperscript{1}\\
	\textsuperscript{1}New York University\quad \textsuperscript{2}Princeton University\\
	\texttt{\{zw3508,ql518\}@nyu.edu, \{jasonlee\}@princeton.edu}
}
\begin{document}

\maketitle 
\begin{abstract}

Understanding when and how much a model gradient leaks information about the training sample is an important question in privacy. In this paper, we present a surprising result: even without training or memorizing the data, we can fully reconstruct the training samples from a single gradient query at a randomly chosen parameter value. 
We prove the identifiability of the training data under mild conditions: with shallow or deep neural networks and a wide range of activation functions. We also present a statistically and computationally efficient algorithm based on tensor decomposition to reconstruct the training data. As a provable attack that reveals sensitive training data, our findings suggest potential severe threats to privacy, especially in federated learning.   
	
\end{abstract}
\section{INTRODUCTION}

It is essential to understand when and how much a model gradient leaks information about the training data. Such a problem poses a major concern especially in federated learning.  Federated learning, where each device can only access its own local training data, has become increasingly popular for training private data in recent years~\citep{brisimi2018federated,mcmahan2017communication}. The defining trait of federated learning is aggregating information without revealing sensitive information in the underlying data. 
Therefore serious privacy concerns arise when one can recover private information from the training procedure. 


Recent years have seen increasingly many studies on reconstructing sensitive data from a model or its gradient updates \citep{deepleak,
	idlg,howeasy,SAPAG,CPL,cafe,howeasy,yin2021see}. This line of work brings about some \textbf{empirical} evidence that the current frameworks in federated learning can be vulnerable to privacy attacks. However, theoretical understanding of this vulnerability remains limited. To fill this gap, we raise the following questions: 

\paragraph{Question 1:}  {\em Is data leakage due to memorization in the training process? In other words, do we need to train the model sufficiently to recover the training samples?}

Many studies have concluded that data reconstruction should be conducted after training the neural network, as the training process leaks more information and helps the model memorize the training data. In this paper, we show the opposite: for a broad class of neural networks architectures, one can reconstruct the samples from a prescribed (random) model's gradient alone \textit{without any training}.

\paragraph{Question 2:} {\em When is the model gradient alone sufficient to identify the training samples (without prior knowledge)? }

Prior work conjectured that a gradient is insufficient to reconstruct the underlying data without prior knowledge \citep{generative}. This conjecture is motivated by some empirical failures in privacy attacks but lacks a sound theoretical foundation. 

Failures in privacy attacks could happen for two different reasons. First, it is possible that the existing optimization algorithms cannot solve the nonconvex reconstruction loss in polynomial time. Second, statistically the gradient alone is insufficient to identify the training samples. Most existing studies have assumed the second case and asserted the need for prior knowledge in privacy attacks \citep{howeasy,yin2021see,generative}. We show in this paper that one can recover the training samples from the gradient alone as long as the model is moderately wide. 

Our paper provides theoretical insights into reconstructing training samples from model gradient (i.e., gradient inversion). Specifically, we have the following contributions:

\begin{itemize}
	\item We show that one gradient query is sufficient to identify the training sample for a broad class of models. Our design applies to fully-connected neural networks with two or more layers and works with most common activation functions, including (Leaky) ReLU, tanh, and sigmoid functions. 
	\item We introduce a statistically and computationally efficient algorithm based on power iteration to reconstruct the training samples. Under some natural assumptions, we show that one can accurately reconstruct both the input and labels when the neural network is $\tilde \Omega(d)$-dimensional wide\footnote{Some polylog factors are hidden.} . 
\end{itemize}

\section{Related Work}
\paragraph{Federated Learning. }
\cite{mcmahan2017communication} proposed the notion of federated learning that the training data is distributed among clients and the shared model is trained by aggregating the updates computed locally. In this way, the central server cannot directly access  the data from clients so is considered safer. Recent works of federated learning improved optimization algorithms \citep{konecny2015federated,konecny2016federated} and communication methods between clients and central server \citep{konecny2016federated2} and tackled related statistical challenges \citep{simith2017federated,zhao2018federated}.
\paragraph{Gradient Inversion. }
Previous works on gradient inversion studied different methods of reconstructing private training data from shared gradients, where a series of optimization-based methods is known as gradient matching. \cite{deepleak} trained minimized the difference between the dummy gradient and true gradient and \cite{idlg} additionally recovered ground truth labels by analyzing the signs of gradients. With a well designed loss function~\citep{howeasy,SAPAG}, initialization~\citep{CPL,cafe}, and image prior regularization~\citep{howeasy,yin2021see}, gradient matching can 
succeed with deeper neural networks, larger batch sizes and higher resolution images.

Furthermore, various of works improved gradient matching method from different perspectives. \cite{huang2021evaluating} relaxed the strong assumption that batch normalization statistics are known. \cite{balunovic2022bayesian} included these optimization-based gradient inversion attacks into an approximation Bayesian adversarial framework. \cite{generative} optimized the difference of gradient on latent space by training a generative model as image prior. \cite{gradvit} generalized gradient matching method to vision transformers. Most of the empirical results in gradient matching indicated that gradient matching alone is insufficient to reconstruct private data, contrary to our theoretical results.

A different framework of gradient inversion is based on an observation by \cite{phong2018privacy} under single neuron setting that $x_k=\nabla_{ W_k}/\nabla_{ b}$, where $b$ is the bias, $x_k$ and $W_k$ is the $k$-th coordinate of the input and the weight respectively.
\cite{Fan2020rethinking} generalized the property to fully connected neural networks and reconstructed private data by solving a noisy linear system. By sequentially constructing the relationship of input and gradient from the output layer to the first layer, \cite{zhu2021rgap,chen2021understanding} proposed methods recovering data from convolutional layers. This series of attacks recover private data with explicit forms but the result may not be unique and can only deal with single input cases.

\paragraph{Tensor Decomposition.}
The notion of tensor decomposition is proposed by \cite{hitchcock1927expression,hitchcock1928multiple} and one instance is known as CP decomposition \citep{Hars1970foundations, carr1970analysis,kiers2000towards}, where a tensor is decomposed into a sum of component rank-1 tensors. For a tensor with higher order, its CP decomposition is often unique \citep{kruskal1977three}. However, solving CP decomposition problem of a tensor is generally NP-hard \citep{haastad1989tensor, hillar2013most} so some restrictions are necessary. A method of solving CP decomposition problem is known as tensor power iteration, which is generalized from matrix power iteration. Some works analyzed this method under orthogonal assumptions \citep{anandkumar2014tensor,wang2015fast,song2016sublinear} or over-complete settings \citep{mcmahan2017communication}.

\section{Preliminaries}
In this section, we formally introduce the problem of reconstructing training data from model gradient, also referred to as the gradient inversion problem~\citep{howeasy,yin2021see,generative}. 

\paragraph{Notation:} We denote lower case symbol $x$ as scalar, bold lower case letter $\bx$ as vector, capital letter $X$ as matrices, and bold capital letter $\bT$ as higher-order tensors. When there is no ambiguity, we also use capital letters (like $Z$) for random variables.  

We use $\sigma$ or $\sigma_k$ to denote the activation function $\sigma:\R\rightarrow\R$. When there is no ambiguity, we also overload same notation for $\sigma:\R^m\rightarrow \R^m$ where the activation is applied coordinate-wise. When $\sigma$ is ReLU it maps $x$ to $(x)_+$ which is $x$ if $x\geq 0$ and 0 otherwise. LeakyReLU: $x\rightarrow (x)_+-0.01(-x)_+$. Tanh activation: $x\rightarrow (e^{2x}-1)/(e^{2x}+1)$, and sigmoid: $x\rightarrow 1/(1+e^{-x})$.

For integer $n$, $[n]:=\{1,2,\cdots n\}$. For vectors we use $\|\cdot\|$ or $\|\cdot\|_2$ to denote its $\ell_2$ norm. For matrices they stand for the spectral norm. We use $\otimes$ to denote tensor product. For clean presentation, in the main paper we use $\tilde O$, $\tilde \Theta$ or $\tilde\Omega$ to hide universal constants, polynomial factors in batch size $B$ and polylog factors in dimension $d$, error $\varepsilon$, total round number $T$ or failure rate $\delta$.

\paragraph{Setup:} 
Consider the supervised learning problem, where we train a neural network $f(\cdot;\Theta):\bx\in\R^d \rightarrow f(\bx;\Theta)\in \R$ through the loss function: 
$$  \min_{\Theta} \sum_{(\bx,y)\in \cD} \ell(f(\bx;\Theta),y). $$
Here $\ell$ is the loss function (we focus on $\ell_2$ loss throughout the paper), and $\cD$ is the dataset of input $\bx\in\R^d$ and label $y\in \R$. 

We consider the setting of federated learning where each client keeps the privacy of their local data. In each iteration between the central server and a client (user machine), each node reports the average of gradient of $\ell(f(\bx;\Theta),y)$ at an unknown batch of their own data $S=\{(\bx_1,y_1),(\bx_2,y_2),\cdots, (\bx_B,y_B)\}$, i.e,
$$G:=\frac{1}{B}\nabla_{\Theta}\sum_{i=1}^B \ell(f(\bx_i,\Theta),y_i). $$ 
Since the batch of samples to be used is determined by the client, the central server cannot ask to query the gradient at the exact same batch of data for the second time. Therefore, our gradient inversion task is as follows: we want to recover the unknown training data $\bx_1,\bx_2,\dots,\bx_B$ with batch size $B$\footnote{The batch size is usually small and can be considered sublinear in $d$.} from the gradient \textbf{queried once} at a model $\Theta$, where the model $\Theta$ and the loss function $\ell$ is known. Since it is known that the labels can be easily recovered by observing the gradient at the last layer \citep{yin2021see,idlg}, most prior work focused on reconstructing the input samples $\bx_i,i\in[B]$.

With the discussed problem, it is straightforward to design the following objective. Indeed almost all the prior work directly solve the following task or its variants:
\begin{align}
\label{eqn:gradient_inversion}
 \min_{ \{\hat\bx_i,\hat y_i\}_{i=1}^B } &  d\left(\frac{1}{B}\sum_{i=1}^B\nabla_{\Theta}\ell(f(\hat\bx_i;\Theta),\hat y_i) ,G \right), 
\end{align}
where $d(\cdot,\cdot)$ is a distance metric of the discrepency between the queried gradient and the estimated gradient (when $\hat \bx_i, y_i$ are now treated as variables to learn). Common choices are $\ell_2$ distance \citep{deepleak,yin2021see} or negative cosine similarity \citep{howeasy}.

 However, such reconstruction loss is nonconvex and over-determined, but consistent nonlinear system. Specifically, since the optimal value of \eqref{eqn:gradient_inversion} is $0$ by design, the above gradient inversion problem is equivalent to solving $M$ equations where $M$ is the dimension of $\Theta$ (or $G$), or namely $B(d+1)$ parameters (the number of unknowns in the training set $S$). Namely, to find the global minimum of \eqref{eqn:gradient_inversion} is equivalent to solving
 \begin{align}
 \label{eqn:system}
 \frac{1}{B}\sum_{i=1}^B\nabla_{\Theta}\ell(f(\hat\bx_i;\Theta),\hat y_i)=G.
 \end{align}
 In order to identify all the training samples, we at least need $B(d+1)\leq M$. In other words, the problem should be over-determined and under-parameterized, which makes the optimization problem even more challenging. (In fact, it is more theoretically sound to solve over-parameterized optimization problems in nonconvex setting~\citep{jacot2018neural,du2019gradient,allen2019convergence}.

Meanwhile, even with a single sample $(B=1)$, the gradient inversion problem in Eqn. \eqref{eqn:gradient_inversion} is NP-complete. (This directly follows from Theorem 1 of \citet{lei2019inverting}.) Therefore we do not hope to solve \eqref{eqn:gradient_inversion} in general without proper constraints on $\Theta$ or its gradient. Instead, we need to carefully choose the $\Theta$ that we query, and design better algorithms to reconstruct the training samples. 

In the next section, instead of directly optimizing over Eqn. \eqref{eqn:gradient_inversion}, we propose some novel algorithms to recover the input data $\bx_i$ and the label $y_i$ by querying at a randomly designed neural network.

\section{Method}

In this section, we introduce the methodology for reconstructing the training samples with fully connected neural networks. As a warm-up, we first focus on two-layer neural networks. We next show that the more general settings for three-layer or deeper neural networks can be reduced to the two-layer cases with similar techniques. 

\subsection{Two-layer Neural Networks}
We first study the case when the model is a 2-layer neural network, denoted by $f(\bx ; \Theta)=\sum_{j=1}^m a_j \sigma\left(\bw_j \cdot \bx\right)$. Here $\Theta=(a_1,\cdots a_m, \bw_1,\cdots \bw_m )$ is a collective way to write the parameters. The input dimension is $d$, and the width of the neural network, or namely the number of hidden nodes is $m$. The objective function in supervised learning is \footnote{Here we exemplify our results using squared loss. It is straightforward to expand our analysis and methodology to other loss functions. }  $$L(\Theta)=\sum_{i=1}^B\left(y_i-f\left(\bx_i ; \Theta\right)\right)^2.$$ 
Notice the gradient with respect to $\ba$ is an $m$-dimensional vector. An interesting phenomenon is that $\nabla_{a_j} f(\bx;\Theta) = \sigma(\bw_j^\top\bx) ,j\in[m] $ only depends on $\bw_j$, and not the other $\bw_k,k\neq j$. Accordingly $\nabla_{a_j}L=\sum_{i=1}^B r_i \nabla_{a_j} f(\bx_i;\Theta)$ only depends on the other $\bw_k,k\neq j$ through each residue $r_i:=f(\bx_i;\Theta)-y_i$. Specifically, we denote the $\nabla_{a_j} L(\Theta)$ as: 
\begin{equation}
\label{gradient}
g_j:=\nabla_{a_j} L(\Theta)=\sum_{i=1}^B r_i \sigma\left(\bw_j ^\top \bx_i\right),
\end{equation}
One important observation is that when the residue $r_i$ is viewed as a constant, $g_j$  becomes a function of $\bw_j$. 

We set $a_j=\frac{1}{m},\forall j\in[m]$, and sample $\bw_j$ independently from standard normal distribution $\cN(0,I)$. With a wide neural network, $r_i=\frac{1}{m}\sum_{k=1}^m \sigma(\bw_k^\top \bx_i) - y_i$ concentrates to a scalar $r_i^*:=\mathbb{E}_{W}[r_i] = \E_{Z\sim \cN(0,\|\bx_i\|^2)}\sigma(Z)-y_i$. When $\sigma$ is odd function like tanh or sigmoid, $r_i^* = -y_i$. 

We defer the formal statements and proof to the next section and the appendix, but with proper concentration, we will have that $g_j = \sum_{i=1}^B r_i^*\sigma(\bw_j^\top \bx_i) + \tilde O(\sqrt{\frac{1}{m}}) $. Now we define the function $g(\bw):=\sum_{i=1}^B r_i^*\sigma(\bw^\top \bx_i)$ be a function on $\bw$. We have $g_j = g(\bw_j) + \tilde O(\sqrt{\frac{1}{m}})$. This is to say by setting different $\bw$ we are able to observe a noisy version of $g(\bw)$, where we have: 
\begin{align}
\nabla g(\bw) = & \sum_{i=1}^B r_i^* \sigma'(\bw^\top \bx_i)\bx_i,\\
\nabla^2 g(\bw) = & \sum_{i=1}^B r_i^* \sigma''(\bw^\top \bx_i)\bx_i\bx_i^\top,\\
\nabla^3 g(\bw) = & \sum_{i=1}^B r_i^* \sigma^{(3)}(\bw^\top \bx_i)\bx_i^{\otimes 3}.
\end{align}
Here $\sigma^{(3)}$ is the third derivative of $\sigma$. This observation suggests that if we are able to estimate $\E_W\nabla^p g(W),$ $ p=1,2,3$ \footnote{It will become necessary to further explore higher order $p=4$ if $\sigma^{(3)}$ is an odd function. This will be explained in Section \ref{sec:theory}.}, we are able to recover the reweighted sum for $\bx_i^{\otimes p}$. Especially when $p=3$, the third order tensor $\E_W\nabla^3 g(W)$ has a unique tensor decomposition which will identify $\{\bx_i\}_{i=1}^B$ when they are independent \citep{kruskal1977three,bhaskara2014uniqueness}.

A natural method to estimate higher-order derivatives is the celebrated Stein's lemma~\citep{stein1981estimation,mamis2022extension}:
\begin{lemma}[Stein's Lemma]
	\label{stein}
	Let $X$ be a standard normal random variable. Then for any function $g$, we have
	\begin{equation}
	\mathbb{E}[g(X)H_p(X)]=\mathbb{E}[g^{(p)}(X)],
	\end{equation}
 if both sides of the equation exists. Here $H_p$ is the $p$th Hermite function and $g^{(p)}$ is the $p$th derivative of $g$.
\end{lemma}
Similarly, when $X$ is vector-valued random variable and $g$ only depends on $X$ through some $\ba^\top X$ ($\ba^\top \sim \cN(0,\|\ba\|^2)$), the $p$-th order polynomials in $H_p$ should be replaced by the $p$-th order (symmetrized) tensor products. Specifically, we will be using $H_3(\bx) = \bx^{\otimes 3}-\bx\Tilde{\otimes} I$, where $\bx\Tilde{\otimes}I(i,j,k)=x_i\delta_{jk}+x_j\delta_{ki}+x_k\delta_{ij}$, where $\delta_{ij}$ is 1 when $i=j$ and 0 otherwise. 

Using Stein's lemma and concentration bounds to (\ref{gradient}) when $a_j=\frac{1}{m}$ and $w\sim \mathcal{N}(0,1)$, we have informally that:
\begin{align}
\notag 
\frac{1}{m}\sum_{j=1}^m g(\bw_j)H_p(\bw_j)\approx & \E_{W\sim \cN(0,I)}[g(W)H_p(W)] \tag{Concentration}\\
=
 \E_W[\nabla^p_Wg(W)] =  & \sum_{i=1}^Br_i^*\mathbb{E}[\sigma^{(p)}(\bw^\top \bx_i)\bx_i^{\otimes p}] \tag{Stein's lemma, and plugging in the gradient of $g$},
\end{align}
Then we can use tensor decomposition to recover $\bx_i$ up to some scaling factors. We present formally our reconstruction procedure in Algorithm \ref{alg:2layer}. 
\begin{algorithm}
	\caption{Two-layer NN: Gradient inversion with tensor decomposition} 
	\label{alg:2layer}	
	\begin{algorithmic}[1]
		\State {\bf \underline{Setup:} } With unknown batch of samples $\{\bx_1,\bx_2,\cdots \bx_B\}$, we have the black box oracle that queries the gradient at transmitted model $\Theta=(\{a_j\}_{j=1}^m, \{\bw_j\}_{j=1}^m)$: $G(\Theta)=(\nabla_{a_i}L(\Theta), \nabla_{\bw_j}L(\Theta))$. 
		\State {\bf \underline{Initialization:} } Set current model $$\Theta: a_j=\frac{1}{m},j\in[m], \text{ and } \bw_j \sim \cN(0,I_d), $$  is sampled from standard normal distribution. Query the gradient $G=(\{g_j\}_{j=1}^m)$ where $g_j=\nabla_{a_j}L(\Theta)$. 
		\State {\bf \underline{Noisy tensor decomposition:}} Set the 3rd order tensor $$\hat \bT:= \sum_{j=1}^m g_jH_3(\bw_j),$$ where $H_3(\bx):=\bx^{\otimes 3}-\bx\tilde\otimes I$. Conduct top-$B$ tensor decomposition of $\hat T$ (e.g. Algorithm 1 in \citep{kuleshov2015tensor}) and get the vectors $\hat \bx_1,\hat\bx_2,\cdots \hat\bx_B$ and recover the weights $\hat{\lambda}_i$. Let $\hat{y}_i=\mathbb{E}_{X\sim \cN(0,\|\hat\bx_i\|^2)}\sigma(X)-\hat{\lambda}_i/\mathbb{E}_{X\sim \cN(0,\|\hat\bx_i\|^2)}\sigma^{(3)}(X)$.
		\State {\bf \underline{Output:}} $\{\hat \bx_1,\hat\bx_2,\cdots \hat\bx_B\}$, $\{\hat y_1,\hat y_2,\cdots \hat y_B\}$.
	\end{algorithmic} 
\end{algorithm}



\subsection{Deep Neural Networks}
\begin{figure*}
	\includegraphics[width=\textwidth]{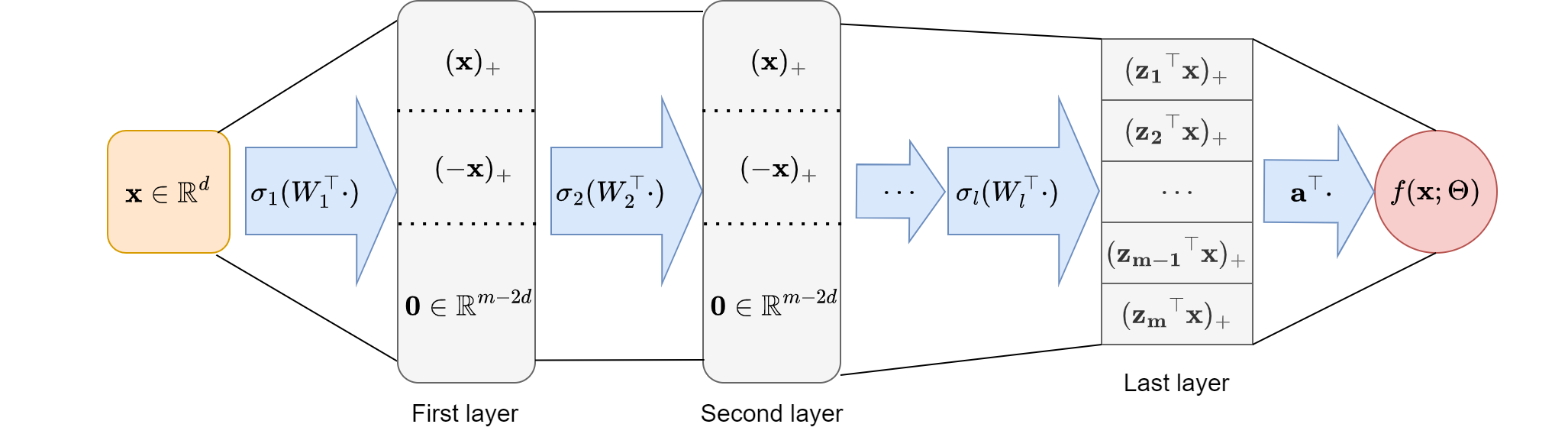}
	\caption{\textbf{Model architecture}: example with ReLU activations to illustrate how multi-layer neural networks can be reduced to the two-layer setting.}
	\label{fig:architecture}
\end{figure*}

\begin{algorithm}
	\caption{Deep neural networks: Gradient inversion with tensor decomposition}
	\label{alg:3layer} 
	\begin{algorithmic}[1]
		\State {\bf \underline{Setup:} } With unknown batch of samples $\{\bx_1,\bx_2,\cdots \bx_B\}$, we have the black box oracle that queries the gradient at transmitted model $$\Theta=(\{a_i\}_{i=1}^m, \{W^{(k)}\}_{k=1}^l),$$ where the slices of $W$ is denoted as $$W^{(k)} = [ \bw_1^{(k)},\bw_2^{(k)},\cdots \bw_m^{(k)} ]. $$ 
		\State {\bf \underline{Initialization:} } Set the model as 
		$$\Theta: a_i=\frac{1}{m},i\in[m], \bw_j^{(l)}=[\bz_j^\top,-\bz_j^\top, 0]^\top, \bz_j\in\R^{2k},$$ where $\bz_j$ is sampled from standard normal distribution.
	 Set $W^{(1)}=[\bar W^{(1)},0]\in \R^{m\times m}$, where $\bar W^{(1)} = [I,-I,0]^\top \in \R^{m\times d}$.
		\For {k=2 to l-1} 
		\State Set $W^{(k)}=[\bar W^{(k)},0]\in \R^{m\times m}$, where $\bar W^{(k)} = [I,0]^\top \in \R^{m\times 2d}$.
		\EndFor 
		\State 
		Query the gradient $G=(\{\bar g_i\}_{i=1}^m)$ where $g_i=\nabla_{a_i}L(\Theta)$. Truncate $ g_i$ as $[\bar g_i,0], g_i\in\R^{D}$.  
		\State {\bf \underline{Noisy tensor decomposition:}} Set the 3rd order tensor $\hat \bT:= \sum_{i=1}^m \bar g_iH_3(\bz_i)$, where $H_3(\bx):=\bx^{\otimes 3}-\bx\Tilde\otimes I$ is the 3rd Hermite polynomial. Conduct top-$B$ tensor decomposition of $\hat T$ (e.g. with Algorithm 1 in \citep{kuleshov2015tensor}) and get the vectors $\hat \bx_1,\hat \bx_2,\cdots \hat \bx_B$ and their corresponding weights $\hat\lambda_i$. Let $\hat y_i = \hat \lambda_i/\lambda$. 
		\If{$\sigma_1$ is LeakyReLU}
		\State 	 $\hat\bx_i \leftarrow \hat\bx_i/1.01$
		\EndIf 
		\State {\bf \underline{Output:}} $\{\hat \bx_1,\hat \bx_2,\cdots \hat\bx_B\}$, $\{\hat y_1,\hat y_2,\cdots \hat y_B \}$.
	\end{algorithmic} 
\end{algorithm} 

Next we move to general deep neural networks. 
Suppose we are working on an $l$-layered deep neural networks, where the function $$f(\bx,\Theta) = \ba^\top \sigma_l(W_l\sigma_{l-1}(W_{l-1}\cdots \sigma_2(W_2\sigma_1(W_1\bx)\cdots ) .$$ 
Here $\ba\in\R^m, W_1\in\R^{m\times d}, W_j\in\R^{m\times m}$ for any $2\leq j\leq l$. 

In this paragraph, we discuss the setting of a malicious parameter server that chooses which model weights to query at and we will later discuss the eavesdropping setting where all weights are at random initialization. When the intermediate layers are wide enough ($m\geq 2d$), we are able to design the first $l-1$ layers to keep the information of each input vector $\bx_i,i\in [B]$. 
Specifically, many activation functions like tanh or sigmoid are bijective. With this type of activations, we will set the weight matrices $W_k,k\in [l-1]$ to identity matrix (concatenated with 0 matrices if the dimension is off) to ensure we don't lose the information of the input. Next, the input of the last layer becomes the output of the bijective function $h_{l-1}(\cdot):=\sigma_{l-1}\circ \cdots \sigma_2\circ \sigma_1(\cdot)$. Therefore, we can view the problem as a same setting of the two-layer case where the input is $h_{l-1}(\bx_i),i\in[B]$. 

For piece-wise linear activations like ReLU:$\bx\rightarrow (\bx)_+$, we will also be able to keep all the information of the input $\bx$ when the number of intermediate hidden nodes $m\geq 2d$. Specifically, we can set $W_1 = [I_d,-I_d,0]^\top$ where $I_d\in\R^{d\times d}$ will ensure that $\sigma(W_1\bx) = [(\bx)_+^\top ,(-\bx)_+^\top ,0]^\top$. For the remaining layers, since the input is now nonnegative, $\sigma(W_j\cdot), 2\leq j\leq l-1$ simply functions as the identity map. Now towards the last but one layer, note that $(\bx)_+ - (-\bx)_+ \equiv \bx$ for any vector $\bx$, and we set $\bw_j=[\bz_j^\top, -\bz_j^\top,0]^\top$. Therefore $\bw_j^\top h_{l-1}(\bx_i)=\bz_j^\top (\bx_i)_+ + (-\bz_j)^\top(-\bx_i)_+ = \bz_j^\top \bx_i$. We can therefore migrate the exact same input vectors to the last but one layer as the same algorithm for the two-layer case. For clear illustration, we present the model architecture for ReLU in Figure \ref{fig:architecture}.

We present more formally the algorithm design in Algorithm \ref{alg:3layer}. For cleaner presentation, we will assume $\sigma_1$ through $\sigma_{l-1}$ are ReLU ($\sigma: \bx\rightarrow (\bx)_+$) or LeakyReLU ($\sigma: \bx\rightarrow (\bx)_+-0.01(-\bx)_+$). We discuss the case for tanh or sigmoid below and also for ReLU with random weights. 

Now we consider the eavesdropping setting. For sigmoid or tanh we consider all the intermediate weight matrices to be at random (Gaussian) initialization, which means with probability $1$ each layer is injective/bijective. 
The only tricky part is that we no longer know the norm of the last but one layer $h_{l-1}(\bx)$. Now suppose we conduct tensor decomposition and get the vectors $\hat \ba_1,\hat \ba_2,\cdots \hat\ba_B$.  We can do a binary search over the correct scaling: find $\alpha_i$ such that $\sigma_1^{-1}(W_1^{\dagger}\cdots \sigma_{l-2}^{-1}(W_{l-2}^{\dagger}\sigma_{l-1}^{-1}(W_{l-1}^{\dagger}\alpha_i \hat \ba_i)\cdots )$ is of norm 1. Suppose we have not done normalization and do not know the norm of the input vectors $\bx_i,i\in [B]$, we can first estimate the correct norm of the last but one layer $h_{l-1}(\bx_i)$. Let's denote $\beta_i:=\|h_{l-1}(\bx_i)\|$. This can be achieved by estimating the weights of the tensor slices for $\bT = \E_{\bw\sim \cN(0,I)} \sigma^{(3)}_{l}(\bw^\top h_{l-1}(\bx_i))h_{l-1}(\bx_i)^{\otimes 3}$. Therefore the weights $\lambda_i = \E_{z\sim\cN(0,\beta_i^2)}\sigma^{(3)}_{l}(z)\beta_i^3$. Since it is only a scalar, we can quickly infer the value of $\beta_i$ from $\lambda_i$.   

For ReLU with random (Gaussian) weights, denote the network by $\ba^\top \sigma_l(f(\bx)) $. By Theorem 5.1 in \citep{paleka2021injectivity}, $f$ is injective with high probability when the layer width is at least $(2l\log l+\Theta(l))d$. If $f^{-1}$ is $L$-Lipschitz on the range of $f$, then it's still possible to identify $\hat\bx_i$ up to error $\|\hat{\bx}_i-\bx_i\|\le \frac{\epsilon}{L}$, where $\epsilon$ is the reconstruction error for two-layer case.

\section{Theoretical Analysis}
\label{sec:theory}
In this section, we will present our main results (informal) with a proof sketch. We defer the more complete proof and detailed dependence (on log factors of dimension $d$, hidden nodes $m$, batch size $B$, failure probability $\delta$) to Appendix \ref{sec:proof}.

\begin{assumption}
\label{assump:2layer}
We make the following assumptions:
    \begin{itemize}
	\item \textbf{Data:} Let data matrix $X:=[\bx_1,\cdots,\bx_B]\in\mathbb{R}^{d\times B}$, we denote the $B$-th singular value by $\pi_{\min}>0$. Training samples are normalized: $\|\bx_i\|=1,\forall i \in [B]$.
	\item \textbf{Activation:} $\sigma$ is 1-Lipschitz and $\E_{z\sim\cN(0,1)}[\sigma''(z)]<\infty$. Let $$k_2=\min\{k\ge 2:|\E_{z\sim \cN(0,1)}[\sigma^{(\alpha)}(z)]|\neq 0\}$$ and $$k_3=\min\{k\ge 3:|\E_{z\sim \cN(0,1)}[\sigma^{(k)}(z)]|\neq 0\}.$$ Then $\nu= |\E_{z\sim \cN(0,1)}[\sigma^{(k_2)}(z)]|$ and $\lambda=| \E_{z\sim \cN(0,1)}[\sigma^{(k_3)}(z)]|$ are not zero. We assume $k_2\le 3$ and $k_3\le 4.$.
\end{itemize}	
\end{assumption}
We note that the data being normalized to $1$ is only a technical assumption to simplify the results. When $\bx_i$ has different norms, the reconstructed accuracy will depend on $\nu_{\min}:=\min_{i}r_i^*\E_{Z\sim \cN(0,\|\bx_i\|^2)}[\sigma''(Z)]$ and $\lambda_{\min}:=\min_{i}r_i^*\E_{Z\sim \cN(0,\|\bx_i\|^2)}[\sigma^{(3)}(Z)]$. $\sigma$ being 1-Lipschitz and having bounded expected second order derivatives is satisfied for the activations we discussed in the paper: (Leaky)ReLU, tanh or sigmoid functions.

The non-degenerate activation, however, is a crucial condition for our algorithm to succeed. For linear or quadratic functions, the third derivative $\sigma^{(3)}$ and the fourth derivative $\sigma^{(4)}$ are 0. Indeed, one can verify that it is in general not possible to recover the individual samples when the activation is linear or quadratic.
For instance, when $\sigma$ is linear, we have $f(\bx;\Theta) = \ba^\top W \bx$, and the gradient with respect to $\ba$ and $W$ are respectively: $$G(\ba)= W (\sum_{i=1}^B r_i \bx_i); G(W)= \ba  (\sum_{i=1}^B r_i \bx_i)^\top.$$ 

Therefore it is only possible to recover a linear combination of all the training samples $\sum_i r_i \bx_i$. Similarly with quadratic activation, we derive the gradient here:
\begin{align*}
\nabla_{a_j} L = \bw_j^\top \bar\Sigma \bw_j; \nabla_{\bw_j}L = 2\bar\Sigma\bw_j,  
\end{align*}
where $\bar\Sigma:=\sum_{i=1}^B r_i\bx_i\bx_i^\top, r_i = f(\bx_i;\Theta) - y_i.$ 

Therefore one can only recover $\bar\Sigma$, a reweighted (weights depending on the residue $r_i$) covariance matrix of all the training samples, or namely the span of the training samples, instead of individual sample $\bx_i$.

\begin{theorem}[Main theorem]
\label{thm:2layer-informal}
	Suppose that $y_i\in\{\pm 1\}$. Under Assumption \ref{assump:2layer}, if we have $B\le \Tilde{O}(d^{1/4})$ and $m \geq \tilde \Omega(\frac{d}{\min\{\nu^2,\lambda^2\}\pi_{\min}^4}) $, then with appropriate tensor decomposition methods, the output of Algorithm \ref{alg:2layer} satisfies:
\begin{align}
\sqrt{\frac{1}{B}\sum_{i=1}^B\|\bx_i-\hat\bx_i\|^2 } \leq &  \frac{1}{\min\{|\nu|,|\lambda|\} \pi_{\min}^2}\tilde O(\sqrt{\frac{d}{m}} )\\
\text{sign}(\hat y_i) = & y_i. 
\end{align}
\end{theorem}
We note that for the uniqueness of tensor decomposition, we only need the samples $\{\bx_i\}$ to be linearly independent (i.e., $\pi_{\min}>0$) \citep{kruskal1977three}. \footnote{\cite{bhaskara2014uniqueness} further proved a robust version for the identifiability.} This is implied by Assumption \ref{assump:2layer}, and suffices the identifiability for the training sets from gradient $G$.  

However, in general tensor decomposition is known to be NP-hard \citep{haastad1989tensor,hillar2013most}, therefore we need some more technical assumptions to constrain the setting in order to derive efficient algorithm. Therefore we adapt the setting of prior work on decomposing tensor by simultaneous matrix diagonalization \citep{kuleshov2015tensor,recovery}, and assume that the minimal singular value of data is non-zero in the main theorem.   

The analysis shows that the neural network is actually very vulnerable to privacy attacks. We are able to recover the images up to an average error $\epsilon$ with mildly overparameterized network when the hidden nodes satisfy $m\gg d/(\min\{\nu^2,\lambda^2\} \pi_{\min}^4 \epsilon^2)$.  

\begin{remark}
	\label{remark:regression}
In the main theorem, for cleaner presentation we considered classification task $y_i\in\{\pm 1\}$. This is not an essential assumption. For regression problems, we can also work on any real-valued and bounded $y_i$. However, the accuracy of tensor power method will depend on $$\kappa = \left|\frac{\lambda_{\max}}{\lambda_{\min}}\right|= \left|\frac{\nu_{\max}}{\nu_{\min}}\right| = \frac{\max_{i\in[B]} |r^*_i|}{\min_{i\in[B]} |r_i^*|}$$ 
If there exists $r_i^*$ too small, we have some tricks to make sure this $\kappa$ is constant. Specifically, suppose $|r_i^*|\leq M$, we can add a large bias term $2M$ in the last layer so that the weights in $\bT$ is in the range of  $[\lambda M, 3\lambda M]$ and the weights in $P$ is in the range of $[\nu M,3\nu M]$, and we can thus ensure $\kappa\leq 3$. In that case, we can instead guarantee 
$$\sqrt{\frac{1}{B}\sum_{i=1}^B |\hat y_i - y_i  |^2 } \leq \frac{1}{\min\{|\nu|,|\lambda|\} \pi_{\min}^2}\tilde O(\sqrt{\frac{d}{m}} ).$$ 
\end{remark}

For deep neural networks, we have the following corollary:
\begin{corollary}
Suppose Assumption \ref{assump:2layer} is satisfied (for $\sigma_l$). Under the same setting for Theorem \ref{thm:2layer-informal}, we have that w.h.p. the output for Algorithm \ref{alg:3layer} satisfies:
\begin{align}
\sqrt{\frac{1}{B}\sum_{i=1}^B\|\bx_i-\hat\bx_i\|^2 } \leq &\frac{1}{\min\{|\nu|,|\lambda|\} \pi_{\min}^2}\tilde O(\sqrt{\frac{d}{m}} ),\\
\text{sign}(\hat y_i) = & y_i. 
\end{align}
\end{corollary}

\subsection{Proof Sketch}

As we demonstrated in the methodology section, the main technique is to estimate $\bT$, where $\bT:=\E[\sum_{i=1}^B r_i^* \sigma^{(3)}(\bw^\top \bx_i) \bx_i^{\otimes 3}] $ and conduct eigendecomposition. We denote $P:=\mathbb{E}[g(\bw)H_2(\bw)]$ and $\hat{P}:=\frac{1}{m}\sum_{j=1}^mg_j(\bw_j)H_2(\bw_j)$, where $H_2(\bx)=\bx\bx^\top-I$. Then the proof of Theorem \ref{thm:2layer-informal} mainly consists of  the concentration bounds of $\hat{P}$ to $P$ in Proposition \ref{matrix} and $\hat\bT$ to $\bT$ in the Proposition \ref{prop:V}, together with perturbation analysis of the tensor method in Proposition \ref{prop:tensorerror}.  

\paragraph{Tensor method.}
There were plenty of results analyzing tensor methods \citep{anandkumar2014tensor,anandkumar2014guaranteed,wang2015fast,wang2016online,song2016sublinear}. We adapt the following result from \citep{recovery} that has a tight dependence on the problem dimension $d$ and few restriction on sample $\{\bx_i\}_{i=1}^B$.

This method first estimated the orthogonal column span $U$ of training samples $\{\bx_i|i\in B\}$ by using power method on $P$, and denoted the estimation by $V$. Then it conducted noisy tensor decomposition to $\bT(V,V,V)$ with Algorithm 1 in \citep{kuleshov2015tensor} and have $\{s_i\bu_i\}_{i=1}^B$ as an estimation of $\{V^\top \bx_i\}_{i=1}^B$, where $s_i\in\{\pm 1\}$ are unknown signs. Finally, by Algorithm 4 in \citep{recovery}, $s_i$, $r_i^*$ and eventually $y_i$ are recovered. The time complexity of this method is $O(Bmd)$ \citep{recovery}.

\begin{proposition}[Adapted from the proof of Theorem 5.6 in \citep{recovery}.]
\label{prop:tensorerror}
    Consider matrix $\hat{P}=P+S$ and tensor $\hat{\bT}=\bT+\bE$ with rank-$B$ decomposition
    $$P=\sum_{i=1}^B\nu_i\bx_i\bx_i^\top,\ \bT=\sum_{i=1}^B\lambda_i\bx_i^{\otimes 3},$$
    where $\bx_i\in \mathbb{R}^d$ satisfying Assumption \ref{assump:2layer}. Let $V$ be the output of Algorithm 3 in \citep{recovery} with input $P$ and $\{s_i\bu_i\}_{i=1}^B$ be the output of Algorithm 1 in \citep{kuleshov2015tensor} with input $\bT(V,V,V)$, where $\{s_i\}$ are unknown signs. Suppose the perturbations satisfy
    $$\|S\|\le\mu,\ \|\bE(V,V,V)\|\le\gamma.$$
    Let $N=\Theta(\log\frac{1}{\epsilon})$ be the iteration numbers of Algorithm 3 in \citep{recovery}, where $\epsilon=\frac{\mu}{\nu_{\min}}$.
    Then w.h.p. we have
    $$\left\|\bx_i-s_i V \bu_i\right\|\le \Tilde{O}(\frac{\mu}{\nu_{\min}\pi_{\min}})+\Tilde{O}(\frac{\kappa\gamma\sqrt{B}}{\lambda_{\min}\pi_{\min}^2}),$$
    where $\kappa=\frac{\lambda_{\max}}{\lambda_{\min}}.$
\end{proposition}

\begin{remark}
\label{rmk:discrete}
    With Assumption \ref{assump:2layer}, the unknown signs $s_i$ and the weights $\lambda_i$ can be recovered by Algorithm 4 in \citep{recovery}. Note that $s_i$ and $\lambda_i$ are discrete so the recovery is exact. Thus, exact $y_i$ can be recovered and the estimation of $\bx_i$ can be represent explicitly.
\end{remark}

To make use the result, what is left is mainly analyzing $\nu_{\min}$, $\lambda_{\min}$ and the weight ratio $$\kappa =  \left|\frac{\lambda_{\max}}{\lambda_{\min}}\right| = \frac{\max_{i\in[B]} |r^*_i| \E[ \sigma^{(3)}(\bw^\top \bx_i) ]}{\min_{i\in[B]} |r_i^*| \E[ \sigma^{(3)}(\bw^\top \bx_i) ]}.$$
Specifically, we have the following claim:
\begin{claim}
	\label{claim:lambda}
Notice the maximum and minimum weights for the tensor slices in $\bT$ are  $$\lambda_{\max}:=\max_{i\in[B]} r_i^*\E[ \sigma^{(3)}(\bw^\top \bx_i) ] = \lambda\max_{i}r_i^*  ,$$ and $$\lambda_{\min}:=\min_{i\in[B]} r_i^*\E[ \sigma^{(3)}(\bw^\top \bx_i) ]  = \lambda\min_{i}r_i^*.$$ For $y_i\in\{\pm 1\}$, and odd activations like sigmoid or tanh,
	$\lambda_{\max}=\lambda_{\min} = \lambda$. 
	Similarly, $\nu_{\max}=\nu\max_ir_i^*$, $\nu_{\min}=\nu\min_ir_i^*$. For $y_i\in\{\pm 1\}$, and activations with odd second order derivative like (Leaky)ReLU, $\nu_{\max}=\nu_{\min}=\nu$.
\end{claim}
This comes from a simple observation from the fact that standard normal distribution is symmetric: $\bw^\top \bx\sim \cN(0,\|\bx\|^2)$ doesn't depend on the direction of $\bx$ but only its norm. Since we have normalized the samples to be norm $1$, that part is invariant to different training sample. We discussed in Remark \ref{remark:regression} how to deal with general $r_i^*$ with its dependence on $\sigma$ and $y_i$. In short, we can easily ensure $\kappa$ to be constant with some small alteration in the designing of the weights.   

In reality, if we deliberately set the norm of some sample $\bx$ to be very small, the coefficient on  the corresponding component of $\bT$ will be very small. This makes the sample hard to learn, which is consistent with our intuition. 

\paragraph{Concentration for matrix and tensor.} We now bound $\mu$ and $\gamma$ in Proposition \ref{prop:tensorerror}. Recall the following notations:
\begin{align}
\hat{P}=&\frac{1}{m}\sum_{j=1}^m\sum_{i=1}^B r_i\sigma(\bw_j^\top\bx_i)(\bw_j\bw_j^\top -I),\\
P=&\E[\sum_{i=1}^B r_i^*\sigma''(\bw^\top \bx_i)\bx_i\bx_i^\top],\\
\hat \bT = & \frac{1}{m}\sum_{j=1}^m\sum_{i=1}^B r_i\sigma(\bw_j^\top \bx_i)(\bw_j^{\otimes 3} - \bw_j\Tilde{\otimes} I) ,\\
\bT = & \E[\sum_{i=1}^B  r_i^*\sigma^{(3)}(\bw^\top \bx_i)\bx_i^{\otimes 3}  ].
\end{align}

\begin{proposition}
\label{matrix}
If $\sigma$ and $\bx_i$ satisfies Assumption \ref{assump:2layer}, $|y_i|\le 1$, then for $\delta\le\frac{2}{d}$ and $m\gtrsim \log(8/\delta)$, we have
\begin{equation}
    \|\hat{P}-P\|
    \le \Tilde{O}(\frac{B\sqrt{d}}{\sqrt{m}})
\end{equation}
with probability $1-\delta$.
\end{proposition}

\begin{proposition}
\label{prop:V}
If $\sigma$ and $\bx_i$ satisfies Assumption \ref{assump:2layer}, $|y_i|\le 1$, and $\|VV^\top-UU^\top\|\le 1/4$, then for $\delta\le\frac{2}{B}$ and $m\gtrsim \log(6/\delta)$
\begin{equation}
    \|\bar{\bT}(V,V,V)-\bT(V,V,V)\|\le\Tilde{O}(\frac{B^{5/2}}{\sqrt{m}})
\end{equation}
with probability $1-\delta$.
\end{proposition}
Here we have omitted the log factor of $\log(BmN/\delta)$ in the inequalities..

\begin{remark}
\label{rmk:tanh}
    For odd activation functions like tanh or sigmoid, $\sigma''$ is an odd function. Due to the symmetry of normal distribution, $\mathbb{E}_{z\sim\mathcal{N}(0,1)}[\sigma''(z)]=0$, which prevents us from using power method. In this case, we instead set
    \begin{align*}
        \hat{P}:=&\frac{1}{m}(\sum_{j=1}^m g_j(\bw_j)H_3(\bw_j))(I,I,\ba) \\
        \approx & \E_{Z\sim \cN(0,1)}[\sigma^{(3)}(Z)](\sum_{i=1}^Br_i^*\bx_i^{\otimes 3})(I,I,\ba),
    \end{align*}
    where $\ba$ is any unit vector. Note that the weight of auxiliary matrix $P$ now is $\nu=\mathbb{E}_{z\sim\mathcal{N}(0,1)}[\sigma^{(3)}]\neq0$. Then all the steps with $P$ in tensor decomposition and their analysis will be similar.
\end{remark}

\begin{remark}
For piecewise linear activations like ReLU or LeakyReLU, note that $\sigma^{(3)}$ is the derivative of Dirac delta function $\delta$, which is odd. Due to the symmetry of normal distribution, $\E_{Z\sim \cN(0,1)}[\sigma^{(3)}(Z)] = 0$, which prevents us from using third order tensor decomposition. In this case, we should instead set  
\begin{align*}
\hat\bT:= &\frac{1}{m}(\sum_{j=1}^m g_j(\bw_j)H_4(\bw_j))(I,I,I,\ba) \\
\approx & \E_{Z\sim \cN(0,1)}[\sigma^{(4)}(Z)](\sum_{i=1}^Br_i^*\bx_i^{\otimes 4})(I,I,I,\ba),
\end{align*}
where $\ba$ is any unit vector.
Then we conduct tensor decomposition to recover $\bx_i$ from $\hat{\bT}$. For instance, the weight in the tensor now becomes $\lambda=\E_{Z\sim \cN(0,1)}[\sigma^{(4)}(Z)]=-\frac{1}{\sqrt{2\pi}}$ for ReLU. All the analysis with $\bT$ still applies in a similar way.  
\end{remark}

\section{Experiments}
\label{sec:exp}
In this section, we present some experimental verification to our theoretical results on synthetic data. Recall that we use the gradient $\nabla_\ba L(\Theta)$ to compute the tensor $\hat{\bT}$. Instead, we can also estimate $\bT$ with the gradient $\nabla_W L(\Theta)$ with respect to the first layer weights $W$ in a similar way (see Appendix \ref{sec:alter}). We mainly use $\nabla_W L(\Theta)$ in our experiments but also make comparison between two methods. We also present the reconstruction result of MNIST in Appendix \ref{sec:mnist}.

\begin{figure*}
    \centering
    \includegraphics[width=7.5cm]{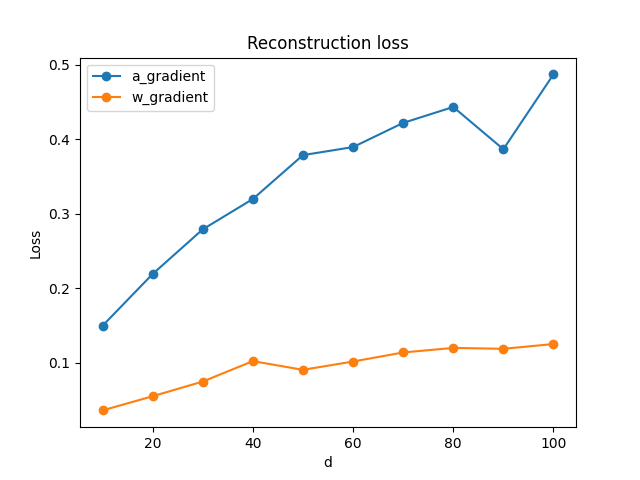}
    \includegraphics[width=7.5cm]{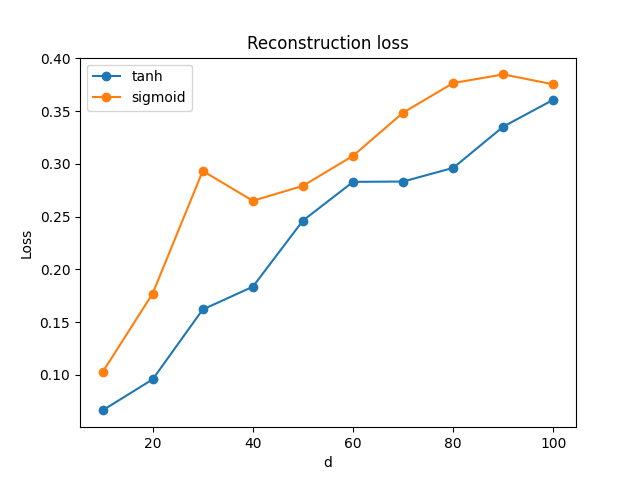}
    \caption{For a two-layer neural network with fixed width $m$, reconstruction loss for different data dimension $d$. \textbf{Left: }using $\nabla_\ba L$ and $\nabla_W L$ when $\sigma(x)=x^2+x^3$; \textbf{Right:} different activation functions tanh and sigmoid using $\nabla_W L$.}
    \label{fig:toy}
\end{figure*} 
We consider a two-layer neural network with fixed width $m=5000$. The data satisfies $B=2$ and $\bx_i=\be_i,$ $i=1,2$. We first set the activation function as $\sigma(x)=x^2+x^3$, a simple example of Assumption \ref{assump:2layer}. When we run Algorithm \ref{alg:2layer} with tensor decomposition method following \cite{recovery}, the reconstruction loss of estimating $\bT$ with $\nabla_\ba L$ and $\nabla_W L$ are shown in the left of Fig. \ref{fig:toy}, where both reconstruction losses are small and using $\nabla_W L$ is better.

For more realistic activation functions, our method can still recover data with small reconstruction loss. The image on the right of Fig. \ref{fig:toy} shows that the construction loss is also small when $\sigma$ is tanh or sigmoid, which aligns with our theoretical results. Here we use the trick in Remark \ref{rmk:tanh}. 

\section{Conclusion and More Discussions}

In this paper, we aim to theoretically investigate the data leakage problem. To the best of our knowledge, this is the first theoretical work to prove that one can reconstruct the training samples from the model gradient. Specifically, we only need mildly overparametrized neural networks, where the width scales \textbf{linearly} (and hidden polylog factors) with the input dimension $d$. In this section, we seek to have more thorough discussions on what can be inferred from our findings, and we believe this area is still wide open for more theoretical investigations. 

\subsection{Discussions} 

\paragraph{On the identifiability of the training samples.}

First, when the training samples are independent, a single gradient alone can identify them (up to some scaling factors). Therefore, if the privacy attacker has unlimited computing power, one can reconstruct the samples. (We make additional assumptions to get a computationally efficient (polynomial-time) reconstructing algorithm. )

Therefore, when prior work finds failure cases in the attack with the gradient, it is more likely that our existing optimization algorithms cannot effectively resolve the non-convex reconstruction loss. Adding prior knowledge to the loss changes the dynamics of the optimization procedure, which might ease the reconstruction procedure.  

\paragraph{On the effect of neural network sizes. }

Based on our results, a wider neural network ensures a more accurate data reconstruction.  
However, a deeper neural network is neither helping nor hurting. This is because, in our current design, the first $l-2$ layers function to preserve the input, while the last two layers generate observations that form a matrix sensing problem. 

From the perspective of equation counting, a deeper network provides more information on identifying the training samples. However, there are also contrary side proofs. When calculating the gradient from a two-layer neural network, especially evident from the linear and quadratic activations, the gradient on the first layer doesn't reveal additional information from the gradient of the second layer. 

Therefore it is not yet clear whether we can benefit from the depth during the reconstruction procedure. It requires further exploration whether this is the caveat of our analysis or the fact that the depth of the network does not play as important a role as the width in privacy attacks.

\paragraph{Distinctions to private algorithms for convex optimization.} 
Our algorithm is similar to a sensing problem as we observe measurements on the input samples, and our target is reconstructing them. However, we only observe the sum of the measurements for individual samples, as the gradient is aggregated with all the training samples. Consequently, linear or quadratic measurements fail the gradient inversion task, as we demonstrated in Section \ref{sec:theory}. 

Meanwhile, since linear functions do not identify individual samples, it is a side proof that the linear and neural network functions are fundamentally different in their vulnerability in privacy attacks. Therefore, although in a very different setting, we also conjecture that previous analyses in differential privacy that worked with convex functions (convex loss on a linear prediction function) \citep{bassily2014private,bassily2019private,altschuler2022privacy} might not generalize to the neural network.   

\paragraph{Discussions on private algorithms.}	
It is common practice to add noise in stochastic convex optimization as privacy-preserving algorithms (especially for differential privacy purposes), namely, by adding noise generated from $Z\sim \cN(0,\sigma^2 I)$ in our observed gradient $G$. Extensive work and analysis demonstrates their ability to keep differential privacy in the convex setting~\citep{bassily2019private,bassily2014private,altschuler2022privacy}. 
However, in our setting, its effect on our reconstruction is limited. In fact, as we observe $g_j = \sum_{i=1}^Br_i\sigma(\bw_j^\top \bx_i) +z_i , z_i\sim \cN(0,\sigma^2) $, our estimated tensor will be shifted with another $\hat \bE:=\sum_{j}z_j H_3(\bw_j)$. One can verify that $\|\bE(I,\ba,\ba)\|\leq \tilde O(\sigma \sqrt{\frac{d}{m} })$ for any vector $\ba$. Therefore as long as $\sigma$ is of constant order, the noisy SGD won't significantly affect (and only changes the constant in) the reconstruction error. 

Therefore, we believe it is more promising to focus on other private algorithms by encoding or perturbing the input samples~\cite{sun2021soteria}. 


\subsection{Future Work} 
Since the theoretical understanding of gradient inversion is still in its nascent phase, our work inevitably has some limitations, and we hope to encourage more future work to fill in the gap. Our work highlights the urgent need to design private algorithms in federated learning. However, here we only focus on the remaining problems of reconstructing sensitive training data. 


From the perspective of parameter counting, for two-layer neural networks, we only need the number of parameters $md$ to be larger than $Bd$, the total dimension of unknown samples. However, we require $m\gg d$ instead of $m\gg B$ in this paper. It is thus important to get either a lower bound or a tighter upper bound to better understand the dependence of hidden nodes $m$ on batch size $B$ and problem dimension $d$. 

Second, it is important to design better attack algorithms to exploit the effect of the depth of neural networks fully. 

Finally but not least, some different neural net architectures will affect the attack. It will be interesting to design the algorithm when the intermediate layers are convolutional instead of fully-connected layers. 
\section*{Acknowledgements}
JDL acknowledges support of the ARO under MURI Award W911NF-11-1-0304,  the Sloan Research Fellowship, NSF CCF 2002272, NSF IIS 2107304,  NSF CIF 2212262, ONR Young Investigator Award, and NSF CAREER Award 2144994.  

\bibliographystyle{abbrvnat}
\bibliography{ref.bib}
%
%




%

%


\appendix 
\section{Proofs}
\label{sec:proof}
\subsection{Concentration Bound for Vectors}
We formalize the concentration bound when $p=1$ in the following propositions.
\begin{prop}
\label{vector}
If $\sigma$ and $\bx_i$ satisfies Assumption \ref{assump:2layer}, $|y_i|\le 1$, then for $\delta\le\frac{6}{d+1}$ and $m\gtrsim \log(6/\delta)$, we have
\begin{equation}
    \left\|\frac{1}{m}\sum_{j=1}^m \sum_{i=1}^B r_i\sigma(\bm{w}_j^\top \bm{x}_i)\bm{w}_j
    -\mathbb{E}\sum_{i=1}^B \Tilde{h}_i'(\bm{w}^\top \bm{x}_i)\bm{x}_i\right\|
    \le \Tilde{O}(\frac{B\sqrt{d}}{\sqrt{m}})
\end{equation}
with probability $1-\delta$, where $\Tilde{h}_i(\bm{w}_j^\top \bm{x}_i)=r_i^*\sigma(\bm{w}_j^\top \bm{x}_i)$.
\end{prop}

The following lemma is crucial to prove the proposition above.
\begin{lemma}[Matrix Bernstein for unbounded matrices; adapted from Lemma B.7 in \citep{recovery}]
\label{bernstein}
Let $\mathcal{Z}$ denote a distribution over $\mathbb{R}^{d_1 \times d_2}$. Let $d=d_1+d_2$. Let $Z_1, Z_2, \cdots, Z_m$ be i.i.d. random matrices sampled from $\mathcal{Z}$. Let $\bar{Z}=\mathbb{E}_{Z \sim \mathcal{Z}}[Z]$ and $\widehat{Z}=\frac{1}{m} \sum_{i=1}^m Z_i$. For parameters $\delta_0 \in(0,1), M=M(\delta_0,m)\ge 0, \nu>0, L>0$,  if the distribution $\mathcal{B}$ satisfies the following four properties,
\begin{center}
$$
\begin{aligned}
        (I)& \quad \mathbb{P}_{Z \sim \mathcal{Z}}\left\{\|Z\| \leq M\right\} \geq 1-\frac{\delta_0}{m}\\
(II)& \quad \max \left(\left\|\underset{Z \sim \mathcal{Z}}{\mathbb{E}}\left[Z Z^{\top}\right]\right\|,\left\|\underset{Z \sim
\mathcal{Z}}{\mathbb{E}}\left[Z^{\top} Z\right]\right\|\right) \leq \nu\\
(III)&  \quad \max _{\|\ba\|=\|\bb\|=1}\left(\underset{Z \sim \mathcal{Z}}{\mathbb{E}}\left[\left(\ba^{\top} Z \bb\right)^2\right]\right)^{1 / 2} \leq L
\end{aligned}
$$
\end{center}

Then we have for any $0<\delta_1<1$, if $\delta_1\le\frac{1}{d}$ and $m\gtrsim \log(1/\delta_1)$,
with probability at least $1-\delta_1-\delta_0$,
$$
\|\widehat{Z}-\bar{Z}\| \lesssim \sqrt{\frac{\log(1/\delta_1)(\nu+\|\bar{Z}\|^2+M\|\bar{Z}\|)+\delta_0L^2}{m}}
$$
\end{lemma}

There are also some useful facts that we need in the proof.
\begin{fact}
\label{norm}
If $\bw$ is a $d$-dimensional standard normal random variable, then
$$
\mathbb{P}\left\{\|\bw\|\ge t\right\}\le2\exp{\left\{-\frac{t^2}{2d}\right\}}.
$$
\end{fact}
\begin{fact}
\label{variance}
If $\bx$ is a $d$-dimensional standard normal random vector, then 
$$\left\|\mathbb{E}_\bw\sigma(\bw^\top\bx)(\bw\bw^\top-I)\right\|\le O(1).$$
\end{fact}
\textit{Proof.} We can assume $\bx=\be_1$ w.l.o.g. and we have that $\mathbb{E}_\bw\sigma(\bw^\top\bx)(\bw\bw^\top-I)$ is diagonal and each element is $O(1)$. \qedsymbol

With the preparation above, we can propose the proof of Proposition \ref{vector} and \ref{matrix}:

\textit{Proof of Proposition \ref{vector}.} By Stein's Lemma, $\mathbb{E}\left[\sum_{i=1}^B \Tilde{h}_i'(\bw^\top \bx_i)\bx_i\right]=\mathbb{E}\left[\sum_{i=1}^B r_i^*\sigma(\bw^\top x_i)w\right]$.
Let $Z_j:=\sum_{i=1}^B r_i\sigma(\bw_j^\top \bx_i)w_j$ and $\Tilde{Z}_j:=\sum_{i=1}^B r_i^*\sigma(\bw_j^\top \bx_i)\bw_j$, then we only need to bound
\begin{equation}
\label{bound}
    \left\|\frac{1}{m}\sum_{j=1}^m Z_j-\mathbb{E}\Tilde{Z}\right\|\le \left\|\frac{1}{m}\sum_{j=1}^m Z_j-\frac{1}{m}\sum_{j=1}^m \Tilde{Z}_j\right\|
    +\left\|\frac{1}{m}\sum_{j=1}^m \Tilde{Z}_j-\mathbb{E}\Tilde{Z}\right\|.
\end{equation}

\paragraph{The first term in Eq. (\ref{bound}).} We first consider $\sigma(\bw_j^\top \bx_i)$. Since $\sigma$ is 1-Lipschitz,
\begin{equation}
    \label{sigma}
\left|\sigma(\bw_j^\top \bx_i)\right|\leq C_\sigma \left|\bw_j^\top \bx_i\right|\le C_\sigma\sqrt{2\log(12B/\delta)}
\end{equation}
with probability $1-\frac{\delta}{6B}$
for some absolute constant $C_\sigma$. Then for another absolute constant $C_\sigma'$, 
\begin{equation}
\label{r}
    \left|r_i-r_i^*\right|=\left|\frac{1}{m}\sum_{j=1}^m\sigma(\bw_j^\top \bx_i)-\mathbb{E}[\sigma(\bw^\top \bx_i)]\right|\le C_\sigma'\sqrt{\frac{\log(12B/\delta)}{m}}
\end{equation}
with probability $1-\frac{\delta}{6B}$. Besides, $|\sigma(\bw_j^\top \bx_i)|\le C_\sigma\sqrt{2\log(12Bm/\delta)}$ with probability $1-\frac{\delta}{6Bm}$ and by Fact \ref{norm}, $\|\bw_j\|\le \sqrt{2d\log(12m/\delta)}$ with probability $1-\frac{\delta}{6m}$. Then by union bound, we have
\begin{equation}
\label{term1}
    \begin{split}
        \left\|\frac{1}{m}\sum_{j=1}^m Z_j-\frac{1}{m}\sum_{j=1}^m \Tilde{Z}_j\right\|
        &=\frac{1}{m}\left\|\sum_{j=1}^m\sum_{i=1}^B(r_i-r_i^*)\sigma(\bw_j^\top \bx_i)\bw_j\right\|\\
        &\le\frac{1}{m}\sum_{i=1}^B\left|r_i-r_i^*\right|\sum_{j=1}^m\left|\sigma(\bw_j^\top \bx_i)\right|\left\|\bw_j\right\|\\
        &\le C_rB\sqrt{\frac{d\log(12B/\delta)\log(12m/\delta)\log(12Bm/\delta)}{m}}
    \end{split}
\end{equation}
with probability $1-\frac{\delta}{2}$, where $C_r$ is an absolute constant.

\paragraph{The second term in Eq. (\ref{bound}).} Since $\bx_i$ and $y_i$ are bounded, $r_i^*=\mathbb{E}[r_i]=y_i-\mathbb{E}[\sigma(\bw^\top \bx_i)]\le|y_i|+C_\sigma\mathbb{E}[|\bw^\top\bx_i|]$ is bounded. Then we check the conditions of Lemma \ref{bernstein}. 

(I) By the proof above, we have
\begin{equation}
\label{M}
    \left\|\Tilde{Z}_j\right\|
\le\sum_{i=1}^B r_i^*\left|\sigma(\bw_j^\top \bx_i)\right|\left\|\bw_j\right\|
\le C_0B\sqrt{d\log(12m/\delta)\log(12Bm/\delta)}
\end{equation}
with probability $1-\frac{\delta}{3m}$.

(II) We have
\begin{equation}
\label{nu}
\begin{split}
    \max\left\{\left\|\mathbb{E}\left[\Tilde{Z}^\top\Tilde{Z}\right]\right\|,\left\|\mathbb{E}\left[\Tilde{Z}\Tilde{Z}^\top\right]\right\|\right\}
&\le \mathbb{E}\left[\left\|\Tilde{Z}\right\|^2\right]
\le B\sum_{i=1}^B\mathbb{E}\left[\left\|r_i^*\sigma\left(\bw_j^\top \bx_i\right)\bw_j\right\|^2\right]\\
&\lesssim B\sum_{i=1}^B\mathbb{E}\left[\sigma\left(\bw_j^\top \bx_i\right)^2\right]^{1/2}\mathbb{E}\left[\left\|\bw_j\right\|^2\right]^{1/2}\\
&\lesssim B^2d.
\end{split}
\end{equation}\

(III) We have 
\begin{equation}
\label{L}
    \max _{\|\ba\|=|b|=1}\left(\underset{Z \sim \mathcal{Z}}{\mathbb{E}}\left[\left(\ba^{\top} Z b\right)^2\right]\right)^{1 / 2} \leq \left(\mathbb{E}\left[\left\|\Tilde{Z}\right\|^2\right]\right)^{\frac{1}{2}}\lesssim B\sqrt{d}.
\end{equation}

Additionally, we also have to bound $\left\|\mathbb{E}\left[\Tilde{Z}\right]\right\|$.
\begin{equation}
\label{mean}
\begin{split}
    \mathbb{E}\left[\Tilde{Z}\right]=\sum_{i=1}^B r_i^*\mathbb{E}\left[\sigma(\bw^\top \bx_i)\bw\right]=\sum_{i=1}^B r_i^*\mathbb{E}\left[\sigma'(\bw^\top\bx_i)\bx_i\right]=\sum_{i=1}^B r_i^*\mathbb{E}_{z\sim\mathcal{N}(0,1)}\left[\sigma'(z)\right]\bx_i
\end{split}
    \end{equation}
by Stein's lemma. Then $\left\|\mathbb{E}\left[\Tilde{Z}\right]\right\|\lesssim \sum_{i=1}^B \|\bx_i\|\lesssim B$

By Eq. (\ref{M})-(\ref{mean}), we can apply Lemma \ref{bernstein}:
\begin{equation}
\label{term2}
    \begin{split}
        \left\|\frac{1}{m}\sum_{j=1}^m \Tilde{Z}_j-\mathbb{E}\Tilde{Z}\right\|
        &\lesssim \sqrt{\frac{\left(\log(6/\delta)\right)\left(B^2d+B^2+B^2\sqrt{d\log(12m/\delta)\log(12Bm/\delta)}+B^2d\right)}{m}}\\
        &\lesssim B\sqrt{\frac{d\log(6/d)\log(12m/\delta)\log(12Bm/\delta)}{m}}
    \end{split}
    \end{equation}
with probability $1-\frac{\delta}{2}$.
    
Putting Eq. (\ref{term1}) and Eq. (\ref{term2}) together, we have
\begin{equation*}
    \left\|\frac{1}{m}\sum_{j=1}^m \sum_{i=1}^B r_i\sigma(\bw_j^\top \bx_i)\bw_j
    -\mathbb{E}\sum_{i=1}^B \Tilde{h}_i'\sigma(\bw^\top \bx_i)\bx_i\right\|
    \le \Tilde{O}(\frac{B\sqrt{d}}{\sqrt{m}})
\end{equation*}
with probability $1-\delta$. \qedsymbol

\subsection{Concentration Bound for $P$}
A lemma is needed for the proof of Proposition \ref{matrix}.
\begin{lemma}
\label{bernbound}
If $\sigma$ is 1-Lipschitz and $\mathbb{E}_{z\sim\mathcal{N}(0,1)}\sigma''(z)<\infty$, $\|\bx\|\le 1$, then for $\delta\le1/d$ and $m\gtrsim\log(2/\delta)$, we have
\begin{equation}
    \left\|\frac{1}{m}\sum_{j=1}^m\sigma(\bw_j^\top\bx)(\bw_j\bw_j^\top-I)-\mathbb{E}\sigma(\bw^\top\bx)(\bw\bw^\top-I)\right\|\lesssim
    \sqrt{\frac{d\log(2/\delta)\log(8m/\delta)}{m}}
\end{equation}
with probability $1-\delta$, where $I$ is the identity matrix.
\end{lemma}
\textit{Proof.} Denote $Z_j=\sigma(\bw_j^\top \bx)(\bw_j\bw_j^\top-I)$. We check the conditions of Lemma \ref{bernstein}.

(I) We first bound the norm of $Z_j$:
\begin{equation}
    \left\|Z_j\right\|\leq
    C_0\left|\sigma(\bw_j^\top\bx)\right|\|\bw_j\|^2\leq
    Cd\log(4m/\delta)
\end{equation}
with probability $1-\frac{\delta}{2m}$, by Fact \ref{norm} and modifying Eq. (\ref{sigma}).

(II) We have
\begin{equation}
    \begin{split}
        \max\left\{\left\|\mathbb{E}\left[Z^\top Z\right]\right\|,\left\|\mathbb{E}\left[Z Z^\top\right]\right\|\right\}
        =\left\|\mathbb{E}\left[Z^2\right]\right\|
        =\left\|\mathbb{E}_{\bw\sim\mathcal{N}(0,I)}\sigma(\bw^\top\bx)^2(\bw\bw^T-I)^2\right\|.
    \end{split}
\end{equation}
We can assume $\bx=\be_1$ w.l.o.g. by the symmetry of normal distribution. Then we have to bound $\|P\|:=\left\|\mathbb{E}\sigma(w_1)^2(\bw\bw^\top-I)^2\right\|$, where we can directly compute that $P$ is a diagonal matrix with positive elements smaller than $(d+1)\max\left\{\mathbb{E}\sigma(w_1)^2w_1^2,\mathbb{E}\sigma(w_1)^2\right\}$. Since $\sigma$ is 1-Lipschitz, $\|B\|\le O(d)$.

(III) For $\max_{\|\ba\|=\|\bb\|=1}(\mathbb{E}(\ba^\top Z \bb)^2)^{1/2}$, it reaches the maximal when $\ba=\bb$ since $Z$ is symmetric. Thus, we have
\begin{equation}
    \mathbb{E}(\ba^\top Z \ba)^2=\mathbb{E}\sigma(\bw^\top\bx)^2(\ba^\top(\bw\bw^\top-I)\ba)^2.
\end{equation}
Similarly, we can assume that $\bx=\be_1$. Then we can compute the expectation:
\begin{equation}
\begin{split}
    \mathbb{E}(\ba^\top Z \ba)^2&=\mathbb{E}\sigma(w_1)^2\left(\sum_{i=1}^d a_i^2(w_i^2-1)+\sum_{i\neq j}a_ia_jw_iw_j\right)^2\\
    &=\mathbb{E}\left[\sum_{i=1}^d \sigma(w_1)^2a_i^4(w_i-1)^2+\sum_{i\neq j}\sigma(w_1)^2a_i^2a_j^2w_i^2w_j^2\right]\\
    &\lesssim \left(\sum_{i=1}^d a_i^2\right)^2=d^2.
\end{split}
\end{equation}
Thus, $\max_{\|\ba\|=\|\bb\|=1}(\mathbb{E}(\ba^\top Z \bb)^2)^{1/2}\le O(d)$.

Moreover, we have to bound $\|\mathbb{E}[Z]\|$, and we have
\begin{equation}
    \mathbb{E}\left[Z\right]=\mathbb{E}\sigma(\bw^\top\bx)(\bw\bw^\top-I)=\mathbb{E}\sigma''(\bw^\top\bx)\bx\bx^\top=\mathbb{E}_{z\sim\mathcal{N}(0,1)}\sigma''(z)\bx\bx^\top
\end{equation}
by Stein's lemma. Then $\|\mathbb{E}[Z]\|\lesssim \|\bx\bx^\top\|\le O(1)$.

By Lemma \ref{bernstein}, we can combine these estimations:
\begin{equation}
    \left\|\frac{1}{m}\sum_{j=1}^m\sigma(\bw_j^\top\bx)(\bw_j\bw_j^\top-I)-\mathbb{E}\sigma(\bw^\top\bx)(\bw\bw^\top-I)\right\|\lesssim
    \sqrt{\frac{\log(2/\delta)(d\log(4m/\delta))}{m}}
\end{equation}
with probability $1-\delta$. \qedsymbol

\textit{Proof of Proposition \ref{matrix}.} Denote $Z_j:=\sum_{i=1}^B r_i\sigma(\bw_j^\top \bx_i)(\bw_j\bw_j^\top-I)$ and $\Tilde{Z}_j:=\sum_{i=1}^B r_i^*\sigma(\bw_j^\top \bx_i)\\
(\bw_j\bw_j^\top-I)$, we have
\begin{equation}
\label{bound2}
    \left\|\frac{1}{m}\sum_{j=1}^m Z_j-\mathbb{E}\Tilde{Z}\right\|\le \left\|\frac{1}{m}\sum_{j=1}^m Z_j-\frac{1}{m}\sum_{j=1}^m \Tilde{Z}_j\right\|
    +\left\|\frac{1}{m}\sum_{j=1}^m \Tilde{Z}_j-\mathbb{E}\Tilde{Z}\right\|.
\end{equation}

\paragraph{The first term in Eq. (\ref{bound2}).} We have
\begin{equation}
\label{term1'}
\begin{split}
    \left\|\frac{1}{m}\sum_{j=1}^m Z_j\right.&\left.-\frac{1}{m}\sum_{j=1}^m \Tilde{Z}_j\right\|\le \sum_{i=1}^B|r_i^*-r_i|\left\|\frac{1}{m}\sum_{j=1}^m\sigma(\bw_j^\top\bx_i)(\bw_j\bw_j^\top-I)\right\|\\
    &\lesssim \sqrt{\frac{\log(8B/\delta)}{m}} \sum_{i=1}^B
   (\|\frac{1}{m}\sum_{j=1}^m\sigma(\bw_j^\top\bx_i)(\bw_j\bw_j^\top-I)-\mathbb{E}\sigma(\bw^\top\bx_i)(\bw\bw^\top-I)\|\\
    &\quad\quad\quad+\left.\left\|\mathbb{E}\sigma(\bw^\top\bx_i)(\bw\bw^\top-I)\right\|\right)\\
    &\lesssim B\sqrt{\frac{d\log(8B/\delta)\log(8B/\delta)\log(16Bm/\delta)}{m}}
\end{split}
\end{equation}
with probability $1-\frac{\delta}{2}$, where the second inequality is by modifying Eq. (\ref{r}) and the last inequality is by Lemma \ref{bernbound} and Fact \ref{variance}.

\paragraph{The second term in Eq. (\ref{bound2}).} By Lemma \ref{bernbound}, we have 
\begin{equation}
    \label{term2'}
    \begin{split}
        \left\|\frac{1}{m}\sum_{j=1}^m \Tilde{Z}_j-\mathbb{E}\Tilde{Z}\right\|&\lesssim \sum_{i=1}^B\left\|\frac{1}{m}\sum_{j=1}^m\sigma(\bw_j^\top\bx_i)(\bw_j\bw_j^\top-I)-\mathbb{E}\sigma(\bw^\top\bx_i)(\bw\bw^\top-I)\right\|\\
        &\lesssim B\sqrt{\frac{d\log(4B/\delta)\log(8Bm/\delta)}{m}}
    \end{split}
\end{equation}
with probability $1-\frac{\delta}{2}$, where the first inequality is because $r^*_i$ is bounded and the second inequality is by Lemma \ref{bernbound}.

Finally we can combine Eq. (\ref{term1'}) and Eq. (\ref{term2'}) and have
\begin{equation*}
    \left\|\frac{1}{m}\sum_{j=1}^m \sum_{i=1}^B r_i\sigma(\bw_j^\top \bx_i)(\bw_j\bw_j^\top-I)
    -\mathbb{E}\sum_{i=1}^B \Tilde{h}_i''(\bw^\top \bx_i)\bx_i\bx_i^\top\right\|
    \le \Tilde{O}(\frac{B\sqrt{d}}{\sqrt{m}})
\end{equation*}
with probability $1-\delta$. \qedsymbol

\subsection{Concentration Bound for \it{T}$(V,V,V)$}
For $U\in \mathbb{R}^{d\times B}$ is the orthogonal column span of $\{\bx_i\}_{i=1}^B$, we assume that the difference between $VV^\top$ and $UU^\top$ is bounded by a constant.

\textit{Proof of Proposition \ref{prop:V}.} We consider $R\in \mathbb{R}^{B\times B^2}$, the flatten along the first dimension of $\bT(V,V,V)$. Since for a symmetric 3rd-order tensor $E$ and its flatten along the first dimension $E^{(1)}$ we have $\|E\|\le\|E^{(1)}\|$, we can bound $\|\hat{R}-R\|$.

Denote $W_j:=\sum_{i=1}^Br_i\sigma(\bw_j^\top\bx_i)(\bw_j^{\otimes3}-3\bw_j\otimes I)$, $\Tilde{W}_j:=\sum_{i=1}^Br_i^*\sigma(\bw_j^\top\bx_i)(\bw_j^{\otimes3}-3\bw_j\otimes I)$, $Z_j:=W_j^{(1)}(V,V,V)$ and $\Tilde{Z}_j:=\Tilde{W}_j^{(1)}(V,V,V)$. Then $\mathbb{E}\Tilde{Z}=\bT^{(1)}(V,V,V)$ by Stein's lemma and we have
\begin{equation}
    \label{bound3}
    \left\|\frac{1}{m}\sum_{j=1}^m Z_j-\mathbb{E}\Tilde{Z}\right\|\le \left\|\frac{1}{m}\sum_{j=1}^m Z_j-\frac{1}{m}\sum_{j=1}^m \Tilde{Z}_j\right\|
    +\left\|\frac{1}{m}\sum_{j=1}^m \Tilde{Z}_j-\mathbb{E}\Tilde{Z}\right\|.
\end{equation}

\paragraph{The first term in Eq. (\ref{bound3}).} Note that $V^\top \bw_j\sim \mathcal{N}(0,I_B)$ so we have
\begin{equation}
\label{term1''}
\begin{split}
    \left\|\frac{1}{m}\sum_{j=1}^m Z_j-\frac{1}{m}\sum_{j=1}^m \Tilde{Z}_j\right\|&\lesssim\frac{1}{m}\sum_{i=1}^B|r_i-r_i^*|\sum_{j=1}^m|\sigma(\bw_j^\top\bx_i)|\|V^\top\bw_j\|^3\\
    &\le\Tilde{O}(\frac{B^{5/2}}{\sqrt{m}})
    \end{split}
\end{equation}
with probability $1-\frac{\delta}{2}$, similar to the proof of Proposition \ref{vector}.

\paragraph{The second term in Eq. (\ref{bound3}).} We check the conditions of Lemma \ref{bernstein}.

(I) By the proof above, we have
\begin{equation}
    \left\|\Tilde{Z}_j\right\|\le\sum_{i=1}^Br_i^*|\sigma(\bw_j^\top\bx_i)|\|V^\top\bw_j\|^3\le\Tilde{O}(\frac{B^{5/2}}{\sqrt{m}})
\end{equation}
with probability $1-\frac{\delta}{3m}$.

(II) We have
\begin{equation}
    \max\left\{\left\|\mathbb{E}\left[\Tilde{Z}^\top\Tilde{Z}\right]\right\|,\left\|\mathbb{E}\left[\Tilde{Z}\Tilde{Z}^\top\right]\right\|\right\}\le\mathbb{E}\left[\left\|\Tilde{Z}\right\|^2\right]\lesssim B\sum_{i=1}^B\mathbb{E}\left[\sigma(\bw_j^\top\bx_i)\right]^{1/2}\mathbb{E}\left[\|V^\top\bw_j\|^6\right]^{1/2}\lesssim B^5.
\end{equation}

(III) We have
\begin{equation}
    \max _{\|\ba\|=|\bb|=1}\left(\underset{Z \sim \mathcal{Z}}{\mathbb{E}}\left[\left(\ba^{\top} Z b\right)^2\right]\right)^{1 / 2}\le \left(\mathbb{E}\left[\left\|\Tilde{Z}\right\|^2\right]\right)^{1/2}\lesssim B^{5/2}.
\end{equation}

Additionally, we also have to bound $\left\|\mathbb{E}\left[\Tilde{Z}\right]\right\|$.
\begin{equation}
\begin{split}
        \mathbb{E}[\Tilde{Z}]&=\sum_{i=1}^Br_i^*\left(\mathbb{E}[\sigma^{(3)}(\bw^\top\bx_i)\bx_i^{\otimes 3}]\right)^{(1)}(V,V,V)\\
        &=\sum_{i=1}^B\mathbb{E}_{z\sim\mathcal{N}(0,1)}[\sigma^{(3)}(z)](V^\top\bx_i)\mathrm{vec}\left[(V^\top\bx_i)(V^\top\bx_i)^\top\right]^\top
\end{split}
\end{equation}
by Stein's lemma. Thus, $\left\|\mathbb{E}\left[\Tilde{Z}\right]\right\|\lesssim B\left\|V^\top\bx_i\right\|^3$. Since $\left\|VV^\top-UU^\top\right\|\le1/4$, we have $\left\|VV^\top\bx_i\right\|=O(1)$. 

Therefor, we can apply Lemma \ref{bernstein} and have
\begin{equation}
\label{term2''}
    \left\|\frac{1}{m}\sum_{j=1}^m \Tilde{Z}_j-\mathbb{E}\Tilde{Z}\right\|\le
    \Tilde{O}(\frac{B^{5/2}}{\sqrt{m}}).
\end{equation}
with probability $1-\frac{\delta}{2}$.

Finally, putting Eq. (\ref{term1''}) and Eq. (\ref{term2''}) together:
\begin{equation}
    \|\bar{T}(V,V,V)-T(V,V,V)\|\le\Tilde{O}(\frac{B^{5/2}}{\sqrt{m}})
\end{equation}
with probability $1-\delta$.
\begin{remark}
The bound in Proposition \ref{prop:V} can be improved if we use finer estimations, e.g. methods similar to the proof of Proposition \ref{matrix}. However, the bound is always $\Tilde{O}(\mathrm{poly}(B))$ since the dimension of $V^\top\bw$ is $B$ so it is not a significant improvement.
\end{remark}

\subsection{Proof for Two-layer Neural Networks}

\begin{theorem}[Main theorem 1; Restated from Theorem \ref{thm:2layer-informal}]
	\label{thm:2layer}
		Suppose $y_i\in\{\pm 1\}$. Under Assumption \ref{assump:2layer}, if $B\le\tilde O(d^{1/4})$ and $m \geq \tilde \Omega(\frac{d}{\min\{\nu^2,\lambda^2\}\pi_{\min}^4}) $, the output of Algorithm \ref{alg:2layer} satisfies:
	\begin{align}
	\sqrt{\frac{1}{B}\sum_{i=1}^B\|\bx_i-\hat\bx_i\|^2 } \leq & \tilde O(\frac{\sqrt{d}}{\min\{|\nu|,|\lambda|\}\pi_{\min}^2\sqrt{m}})\\
	\text{sign}(\hat y_i) = & y_i. 
	\end{align}
\end{theorem}

The poof for Theorem \ref{thm:2layer} directly applies Proposition \ref{prop:tensorerror} and Remark \ref{rmk:discrete}. The Proposition \ref{matrix} assigns $\mu=\tilde O(B{\frac{B^{5/2}}{\sqrt{m}}})$ and Proposition \ref{prop:V} assigns $\gamma=\tilde O(\sqrt{\frac{d}{m}})$. Claim \ref{claim:lambda} helps determine that $\kappa$ is constant. We have also demonstrated how to work with small $r_i^*$ in Remark \ref{remark:regression}. After plugging in the values used in Theorem \ref{prop:tensorerror} we natually get the results shown in Theorem \ref{thm:2layer}.

\section{Alternative Method}
\label{sec:alter}
In this section, we will introduce an alternative method to compute the estimated tensor $\hat{\bT}$. We denote $\hat{g}_j$ as:
$$
\hat{g}_j:=\nabla_{w_j} L(\Theta)=\sum_{i=1}^B r_i a_j\sigma'\left(\bw_j ^\top \bx_i\right)\bx_i.
$$
Let $a_j=\frac{1}{m},\forall j\in[m]$ and $\bw_j\in\mathcal{N}(0,1)$, by Stein's lemma, we have
\begin{align*}
    \notag
    \bT_1&:=\sum_{j=1}^m\hat{g}_j H_2(\bw_j)=\frac{1}{m}\sum_{i=1}^B r_i^* \bx_i\otimes\left[\sum_{j=1}^m\sigma'\left(\bw_j ^\top \bx_i\right)\left(\bw_j\otimes\w_j-I\right)\right]\\
    &\approx \sum_{i=1}^B r_i^*\bx_i\otimes\mathbb{E}\left[\sigma'\left(\bw_j ^\top \bx_i\right)\left(\bw_j\otimes\w_j-I\right)\right]\\
    &=\sum_{i=1}^Br_i^* \mathbb{E}\left[\sigma^{(3)}(\bw^\top \bx_i)\right]\bx_i^{\otimes 3}=\bT.
\end{align*}
Let $\bT_2$ and $\bT_3$ be tensors such that $\bT_2(i,j,k)=\bT_1(k,i,j)$ and $\bT_3(i,j,k)=\bT_1(j,k,i)$. Then $\hat{\bT}:=\frac{\bT_1+\bT_2+\bT_3}{3}\approx\bT$ and is symmetric. As shown in Section \ref{sec:exp}, estimating $\bT$ by gradient with respect to $W$ is empirically better than original method. Therefore, we use this estimation of $\bT$ in all experiments (except for the experiment comparing two methods).

\section{Experiment Dertails}
In our experiments in Section \ref{sec:exp}, we reconstruct training data from two-layer neural networks with width $m=5000$. The synthetic training data is $\bx_i=\be_i,i=1,2$ with batch size $B=2$ and labels $y_1=1, y_2=-1$. A bias term $M=30$ is added to $r_i$ in experiments following Remark \ref{remark:regression}. We run Algorithm \ref{alg:2layer} with tensor method in \citep{recovery}. For noisy tensor decomposition conducted on $\bT(V,V,V)$, the number of random projections of Algorithm 1 in \citep{kuleshov2015tensor} is fixed as $L=100$.

\section{Additional Experiment Results}
\label{sec:mnist}
\begin{figure}[H]
    \centering
    \includegraphics[width=7.5cm]{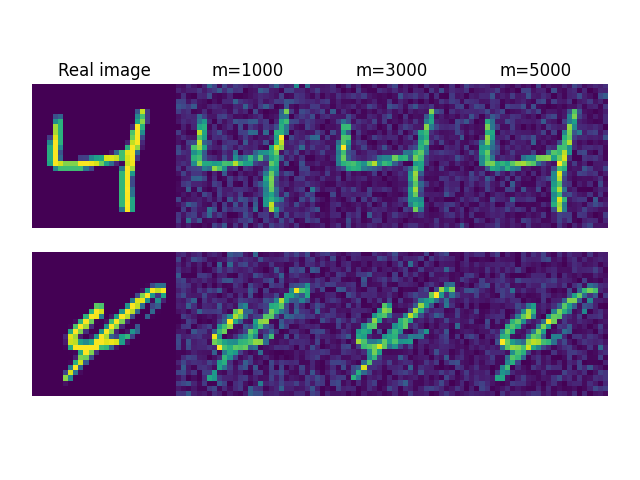}
    \includegraphics[width=7.5cm]{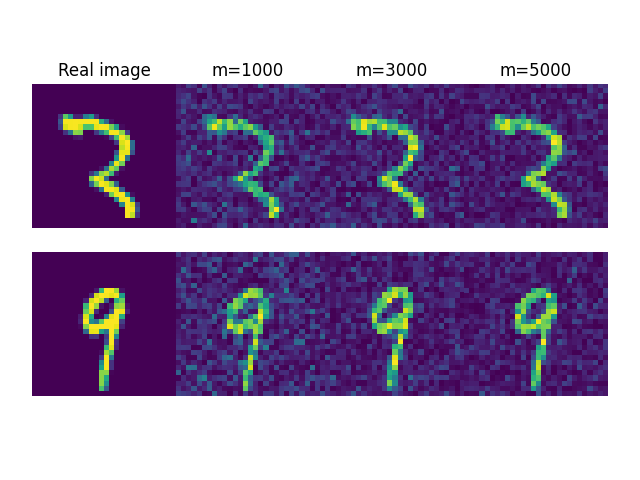}
    \caption{Reconstruction images of MNIST with different network width $m$. \textbf{Left: }images with a same label; \textbf{Right: }images with different labels. }
    \label{fig:mnist}
\end{figure}

In this section , we will present some additional empirical results on real data. We conduct Algorithm \ref{alg:2layer} to reconstruct MNIST images from a binary classification problem with square loss. For simplicity, we set batch size $B=2$ and activation function $\sigma(x)=x^2+x^3$. Different network width $m$ are used in this experiment, when data dimension $d$ is fixed as 784. We use other settings same as the experiments on synthetic data. The results here are the reconstruction images where the two images in the batch have same or different labels (Fig. \ref{fig:mnist}).

\end{document}